\documentclass[a4paper]{article}

\usepackage{INTERSPEECH_v2}

\title{Phone-aware Neural Language Identification}
\name{Zhiyuan Tang, Dong Wang, Yixiang Chen, Ying Shi, Lantian Li}
\address{Center for Speech and Language Technologies (CSLT), Tsinghua University}
\email{Corresponding author:wangdong99@mails.tsinghua.edu.cn}

\begin{document}

\maketitle
\begin{abstract}

 Pure acoustic neural models, particularly the LSTM-RNN model, have shown
 great potential in language identification (LID). However, the phonetic
 information has been largely overlooked by most of existing
 neural LID models, although this information has been used in
 the conventional phonetic LID systems with a great success.
 We present a phone-aware neural LID architecture, which
 is a deep LSTM-RNN LID system but accepts output from an RNN-based
 ASR system. By utilizing the phonetic knowledge, the LID performance can be
 significantly improved. Interestingly, even if the test language is not
 involved in the ASR training, the phonetic knowledge still presents a
 large contribution. Our experiments conducted on four languages within the
 Babel corpus demonstrated that the phone-aware approach is highly effective.

  \end{abstract}

\section{Introduction}
\label{sec:intro}

 Language identification (LID) lends itself to a wide range of applications, e.g., mix-lingual (code-switching)
 speech recognition. Early methods are based on statistical models of phonetic or acoustic units~\cite{lamel1994language,zissman1996comparison,li2007vector}. Recent methods are
 based on probabilistic acoustic modeling, among which the i-vector model is perhaps the most successful~\cite{Najim2011lang,martinez2011language}.

 Recently, deep neural models have attracted much attention in LID. Lopez-Moreno et al.~\cite{lopez2014automatic} proposed a
 DNN-based approach which uses a DNN to discriminate different languages at the frame-level, and the
 language posteriors of an utterance are generated by a simple average of all the frame-level posteriors of the utterance.
 An RNN-based approach was later proposed by Gonzalez-Dominguez et al.~\cite{gonzalez2014automatic}, and better
 performance was obtained with much less parameters compared to the DNN-based model. Due to the
 advantage in temporal modeling, the RNN approach
 has been followed by a number of researchers, e.g.,~\cite{gelly2016divide,zazo2016language}.
 Other neural model structures were
 also investigated, e.g., CNN~\cite{lozano2015end,jin2016lid} and TDNN~\cite{kotov2016language,garcia2016stacked}.
 Compared to the i-vector approach that is based on a probabilistic model, these pure neural methods show clear advantage in short
 utterances (e.g., 2$\sim$3 seconds)~\cite{lopez2014automatic,gonzalez2014automatic,zazo2016language}. The main advantage of the neural-based methods,
 compared to the i-vector model, is that they are discriminative and can learn complex decision bounds
 between languages, provided that sufficient data is provided. Moreover, the power of feature learning
 associated with deep neural nets often provides better robustness against noise and speaker variation,
 which is highly important for LID.

 Deep neural models are also used in a hybrid way, i.e., to generate features~\cite{song2013vector} or
 alignment~\cite{ferrer2016study,tian2016investigation} for an i-vector model. In this case, the
 phonetic DNN/RNN model is trained for phone discrimination as in automatic speech recognition (ASR).
 This model is then used to produce bottle-neck features or acoustic alignment for constructing the i-vector model.
 By using the phonetic information that is directly related to LID, the i-vector model can be consistently improved.

 The above two-approaches have their own disadvantages. For the pure neural approach,
 the entire system relies on raw features, ignoring any phonetic information that is known
 to be important from the beginning of LID research~\cite{zissman1996comparison}; for the hybrid system,
 it is still based on a probabilistic model that (1)
 involves a strong Gaussian assumption that is not suitable for dealing with complex class (here, language) boundaries; (2)
 requires relatively more speech frames to estimate a reliable i-vector, which is not applicable to
 many real applications that require quick identification, e.g., code-switch ASR.

\begin{figure}[h]
  \centering
  \includegraphics[width=\linewidth]{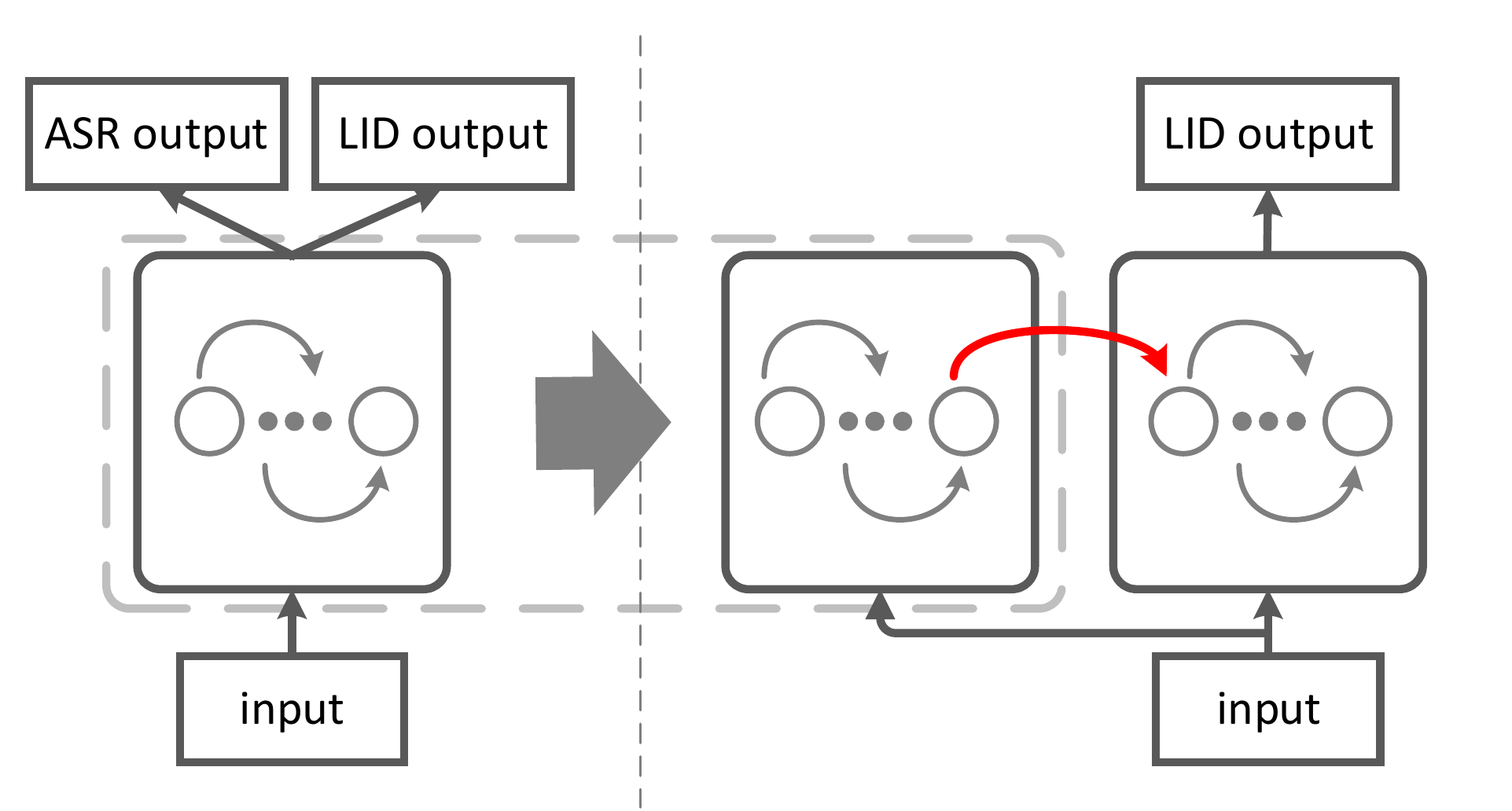}
  \caption{The architecture of phone-aware neural language identification.}
  \label{fig:lid}
\end{figure}

 In this paper, we follow the pure neural model scheme, and enhance the existing models by
 introducing phonetic information as an auxiliary feature. Due to the clear advantage of the LSTM-RNN in both
 ASR and LID, we adopt this model in the study, though the idea of leveraging phonetic information
 is applicable to any neural models.
 The architecture is illustrated in Figure~\ref{fig:lid}, which involves a
 phonetic RNN that is trained to discriminate phones as in ASR and produces phonetic features
 once the training is done, and an LID RNN which receives the phonetic features and uses them together
 with the raw acoustic feature to perform LID. This model has the following properties:

 \begin{itemize}

 \item The phonetic RNN can be trained with flexible objectives. It can be discriminant for phones (as in ASR)
  or for both phones and languages, following the multi-task learning principle~\cite{caruana1997multitask}.

  \item The phonetic RNN can be trained independently from the LID RNN. This means that it can be trained using
  data of any languages that are totally different from the target languages of the LID task. This is
  particularly attractive when the  LID task is to discriminate low-resource languages.

  \item The phonetic feature extraction and propagation is flexible. It can be extracted from any place of the
  phonetic RNN, and can be propagated to any place of the LID RNN.
 \end{itemize}


 This architecture is a reminiscence of the early phonetic recognition and language modeling (PRLM) approach~\cite{matejka2006brno},
 where a phone recognizer
 is used as a front-end to decode phonetic units, followed by a phonetic LM to perform scoring. The two RNNs in our architecture
 can be regarded as corresponding to the phone recognizer and the LM respectively, although the structure is much more
 flexible than the historic model.
 In fact, if the phonetic features are derived from the output layer of the phonetic RNN and are propagated to the input layer of the
 LID RNN, and if the raw feature is omitted, we obtain a PRLM system where the LID RNN is essentially an RNN-based phone
 LM. This architecture was recently studied by Salamea et al.~\cite{salamea2016use}.





The rest of the paper is organized as follows: the model structure is described
in Section~\ref{sec:model} and the experiments are reported in Section~\ref{sec:exp}, followed by
some conclusions and future work in Section~\ref{sec:con}.

\section{Model structure}
\label{sec:model}

We choose the LSTM-RNN as the phonetic RNN and the LID RNN components in the study.
One reason for the choice is that LSTM-RNN has been demonstrated to perform well
in both the pure neural approach~\cite{gonzalez2014automatic}
and the hybrid approach~\cite{tian2016investigation}.
Another reason is that this structure (phonetic feature plus RNN LID) is in accordance with our
motivation to model the phonetic dynamics as in the old PRLM approach~\cite{matejka2006brno}. This section
first describes the LSTM-RNN structure used in the study, and then presents the phone-aware LID system.

\subsection{LSTM structure}

\begin{figure}[h]
  \centering
  \includegraphics[width=\linewidth]{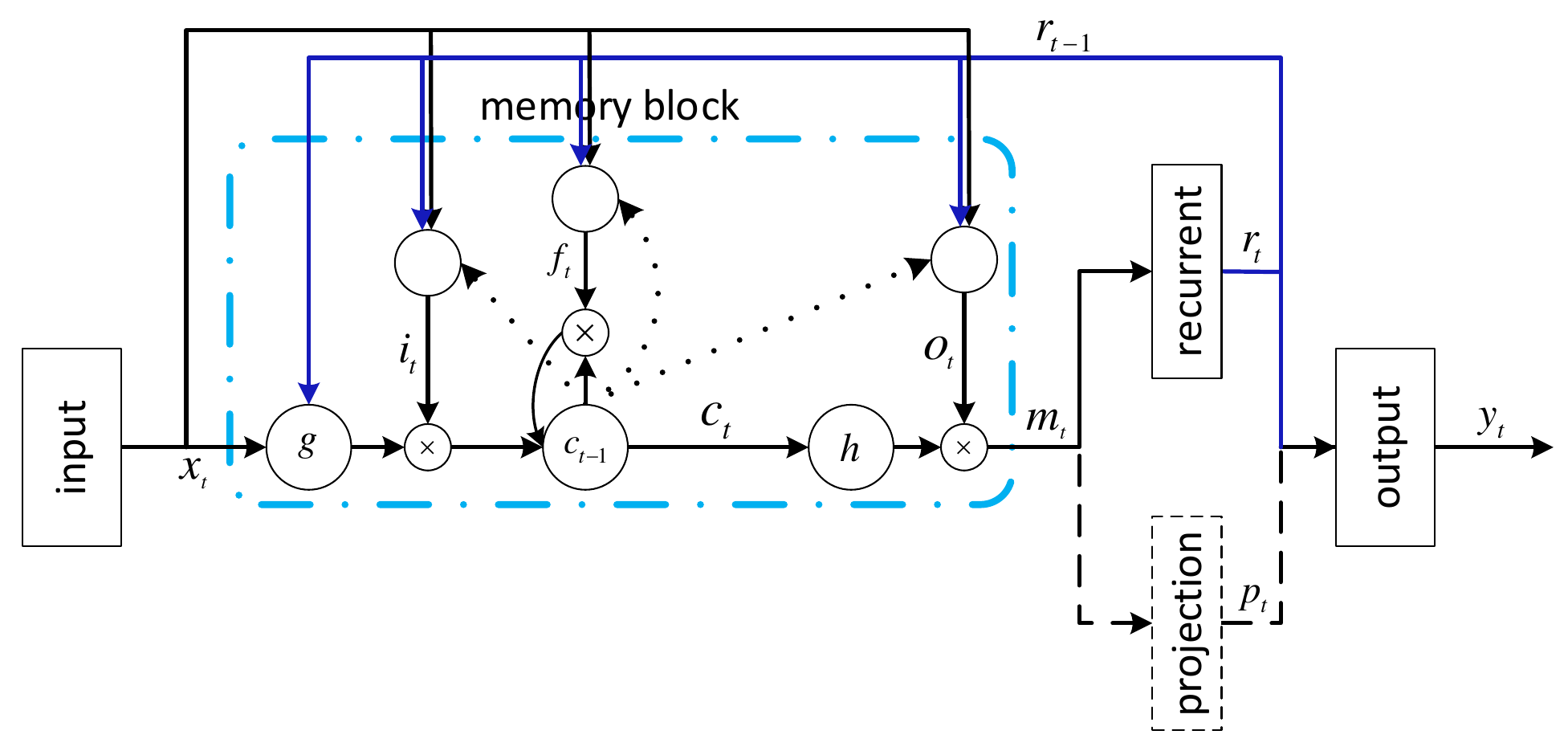}
  \caption{The LSTM model for the study. The picture is reproduced from~\cite{sak2014long}.}
  \label{fig:lstm}
\end{figure}

The LSTM model proposed in~\cite{sak2014long} is used in the study, as shown in Figure~\ref{fig:lstm}.

The associated computation is given as follows:

\begin{eqnarray}
  i_t &=& \sigma(W_{ix}x_{t} + W_{ir}r_{t-1} + W_{ic}c_{t-1} + b_i) \nonumber\\
  f_t &=& \sigma(W_{fx}x_{t} + W_{fr}r_{t-1} + W_{fc}c_{t-1} + b_f) \nonumber\\
  c_t &=& f_t \odot c_{t-1} + i_t \odot g(W_{cx}x_t + W_{cr}r_{t-1} + b_c) \nonumber\\
  o_t &=& \sigma(W_{ox}x_t + W_{or}r_{t-1} + W_{oc}c_t + b_o) \nonumber\\
  m_t &=& o_t \odot h(c_t) \nonumber\\
  r_t &=& W_{rm} m_t \nonumber\\
  p_t &=& W_{pm} m_t \nonumber\\
  y_t &=& W_{yr}r_t + W_{yp}p_t + b_y \nonumber
\end{eqnarray}

\noindent In the above equations, the $W$ terms denote weight matrices and those associated with the cells were constrained to be diagonal in our implementation.
The $b$ terms denote bias vectors. $x_t$ and $y_t$ are the input and output symbols respectively; $i_t$, $f_t$, $o_t$ represent respectively
the input, forget and output gates; $c_t$ is the cell and $m_t$ is the cell output. $r_t$ and $p_t$ are two output components derived from $m_t$, where $r_t$ is recurrent and fed to the next time step, while $p_t$ is not recurrent and contributes to the present output only.
$\sigma(\cdot)$ is the logistic sigmoid function, and $g(\cdot)$ and $h(\cdot)$ are non-linear activation functions, often chosen to be hyperbolic. $\odot$ denotes element-wise multiplication.

\subsection{Phone-aware LID system}

\begin{figure}[h]
  \centering
  \includegraphics[width=0.95\linewidth]{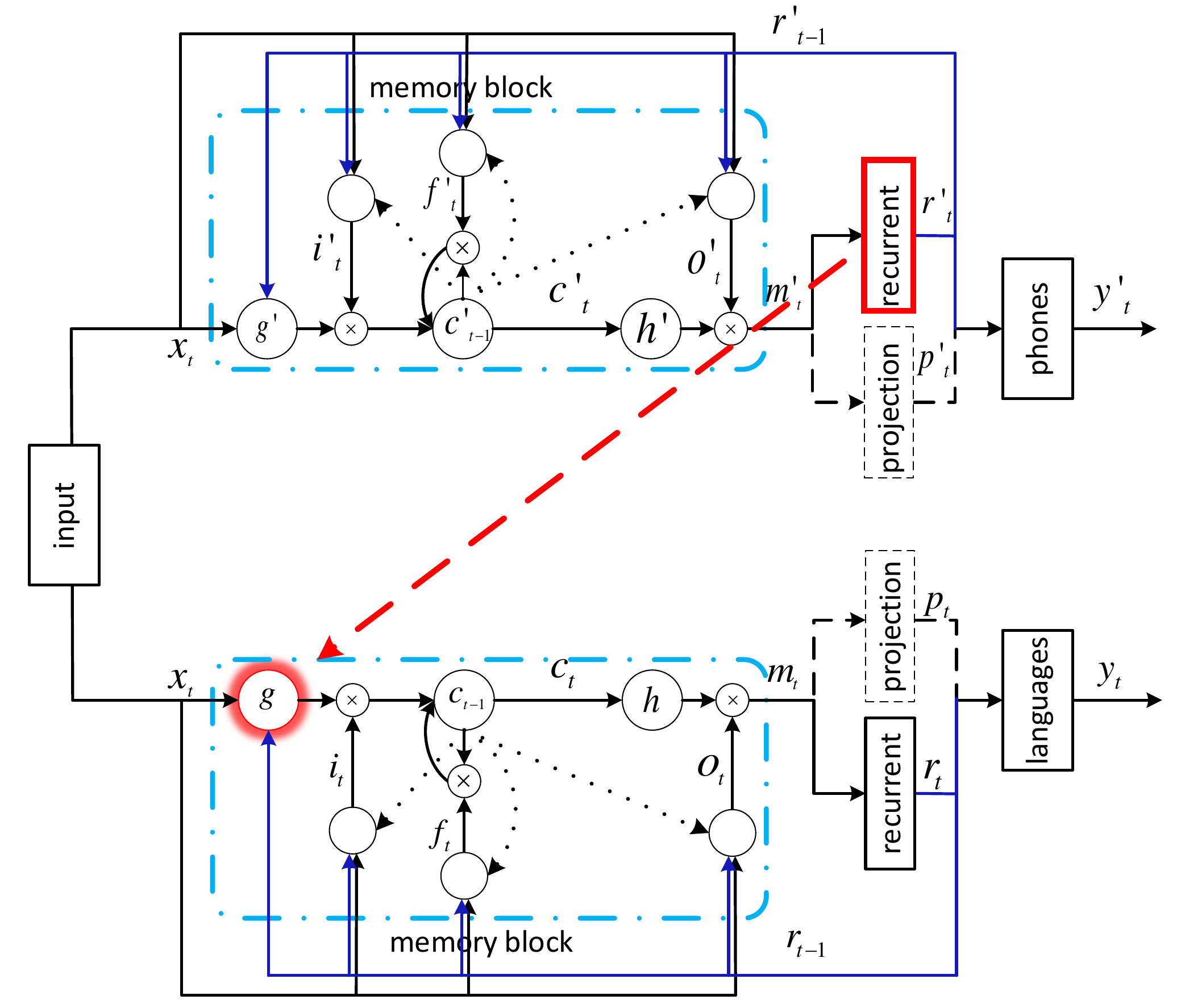}
  \caption{The phone-aware LID system.}
  \label{fig:sys}
\end{figure}

As a preliminary work to demonstrate the concept, we design a simple phone-aware LID system as shown in Figure~\ref{fig:sys},
where both the phonetic RNN and the LID RNN involve a single LSTM layer. Although the phonetic features can be extracted from
any places of the phonetic RNN, we choose to use the output of the recurrent projection layer. Similarly, the receiver of the
phonetic features is also flexible and we will investigate the performance of difference choices. The configure shown in
Figure~\ref{fig:sys} uses the non-linear function $g(\cdot)$ as the receiver. With this configure, most computation of the
LID RNN remains the same, except that the cell value should be updated as follows:

\[
  c_t = f_t \odot c_{t-1} + i_t \odot g(W_{cx}x_t + W_{cr}r_{t-1} + \underline{W'_{cr}r'_{t}} + b_c)
\]

\noindent where $r'_t$ is the phonetic feature propagated from the phonetic RNN.

In this study, all the RNN models use the same configuration that the LSTM layer consists of $1,024$
cells, and the dimensionality of both the recurrent and non-recurrent projections is set to $256$.
The natural stochastic gradient descent (NSGD) algorithm ~\cite{povey2014parallel} is employed to
train the model.

\section{Experiments}
\label{sec:exp}

\subsection{Database and configurations}

The experiments were conducted with the Babel corpus. This corpus was collected
as part of the IARPA (Intelligence Advanced Research Projects Activity) Babel program, with
aim to develop speech technologies for low-resource languages.
In this paper, we chose speech data of four languages from the Babel corpus
to conduct the study: Assamese, Georgian,
Bengali, and Turkish. For each language, a training dataset and a development dataset
were officially provided. Training dataset contains both conversational speech and scripted speech
and development dataset only has conversational speech.
We used the entire training set of each language in the model training, but randomly
selected $2,000$ utterances from the development set of each language to perform the test.
The training data sets from the four languages are as follows:
Assamese\footnote{Language collection release IARPA-babel102b-v0.5a.} $187$ hours,
Georgian\footnote{Language collection release IARPA-babel404b-v1.0a.} $159$ hours,
Bengali\footnote{Language collection release IARPA-babel103b-v0.4b.} $217$ hours, and
Turkish\footnote{Language collection release IARPA-babel105b-v0.5.} $268$ hours.
The average length of the test utterances is $9.50$ seconds, ranging from $0.48$ seconds to $56.78$
seconds.

The raw feature used for the RNN models is $23$-dimensional Fbanks, with a symmetric $2$-frame window to splice neighboring frames.
All the experiments were conducted with Kaldi~\cite{povey2011kaldi}. The default
configurations of the Kaldi WSJ s5 nnet3 recipe were used to train the phonetic RNN
and the LID RNN.

\subsection{Baseline results}
\label{subsec:base_results}

As the first step, we build three RNN baseline systems using the speech data of
two languages: Assamese and Georgian. The three RNN baselines are:
multilingual ASR system (AG-ASR), LID system (AG-LID), ASR-LID multi-task system (AG-MLT).
For the AG-ASR, the phone sets of the two languages are merged and the softmax group
involves all the state targets, which is $3,349$ in our experiment. The ASR performance
in terms of word error rate (WER) is $68.2\%$ and $65.5\%$ for Assamese and Georgian on the whole
development dataset, respectively.
The training
and decoding follow the standard WSJ s5 nnet3 recipe of Kaldi.
For the AG-LID, the output
layer consists of two units, corresponding to the two languages respectively. The training
procedure is similar to the one used for training the AG-ASR model. The AG-MLT model involves two groups
of targets, and the training utilizes the labels of both phones and languages.

The LID for Assamese and Georgian can be conducted by either AG-LID or AG-MLT, using the language posteriors
they produce.
The performance results with these two systems, in terms of $C_{avg}$ and equal error rate (EER), are
shown in Table~\ref{tab:base}. Both the frame-level performance and the
utterance-level performance are reported. For the utterance-level results, the
frame-level posteriors are averaged to produce the utterance-level posterior, with which
the LID is conducted.

\begin{table}[thb!]
  \caption{\label{tab:base}{LID results of the baseline systems on known languages (Assamese and Georgian).}}
  \centerline{
    \begin{tabular}{l|cc|cc}
      \hline
                & \multicolumn{2}{c}{$C_{avg}$}   & \multicolumn{2}{|c}{EER\%} \\
      \hline
      Model        &    Fr. & Utt. & Fr.    &  Utt. \\
      \hline
      AG-LID   &    0.1545    & 0.0905  & 16.20     & 9.20  \\
      AG-MLT  &    0.0822    & 0.0399  & 8.68       & 4.10  \\
      \hline
   \end{tabular}
  }
\end{table}

The results in Table~\ref{tab:base} indicates that both the LID RNN and the multi-task LID RNN are capable of
language discrimination, and the multi-task RNN performs better. This is expected as the ASR targets
can help to regularize the model training and alleviate the impact of  variability factors such as noise and speaker.

\subsection{Phonetic feature}
\label{sec:discrimination}

\begin{figure}[h]
  \centering
  \includegraphics[width=\linewidth]{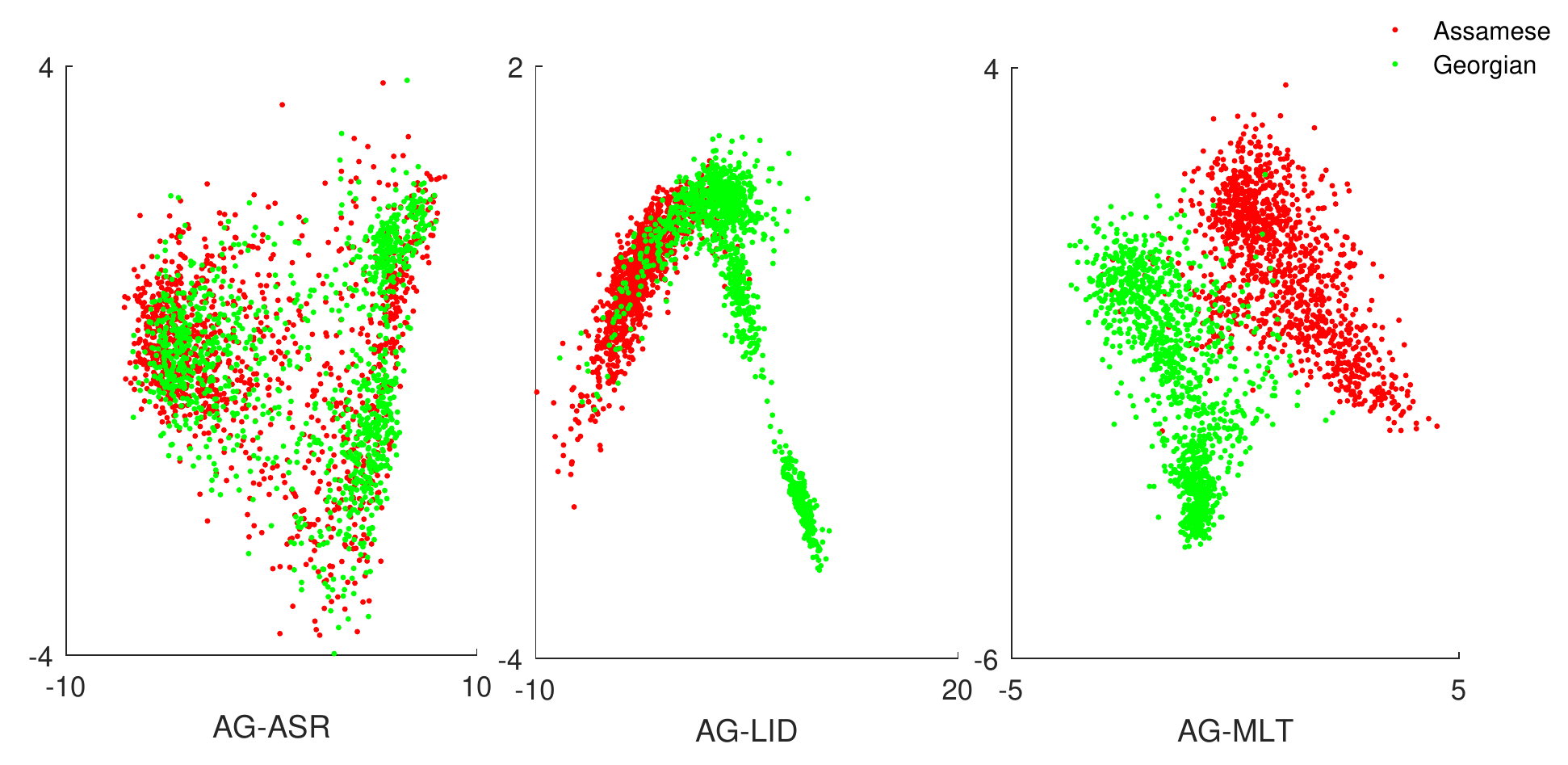}
  \caption{Features of the speech data from Assamese and Georgian, produced by the three phonetic RNNs.}
  \label{fig:visual12}
\end{figure}

\begin{figure}[h]
  \centering
  \includegraphics[width=\linewidth]{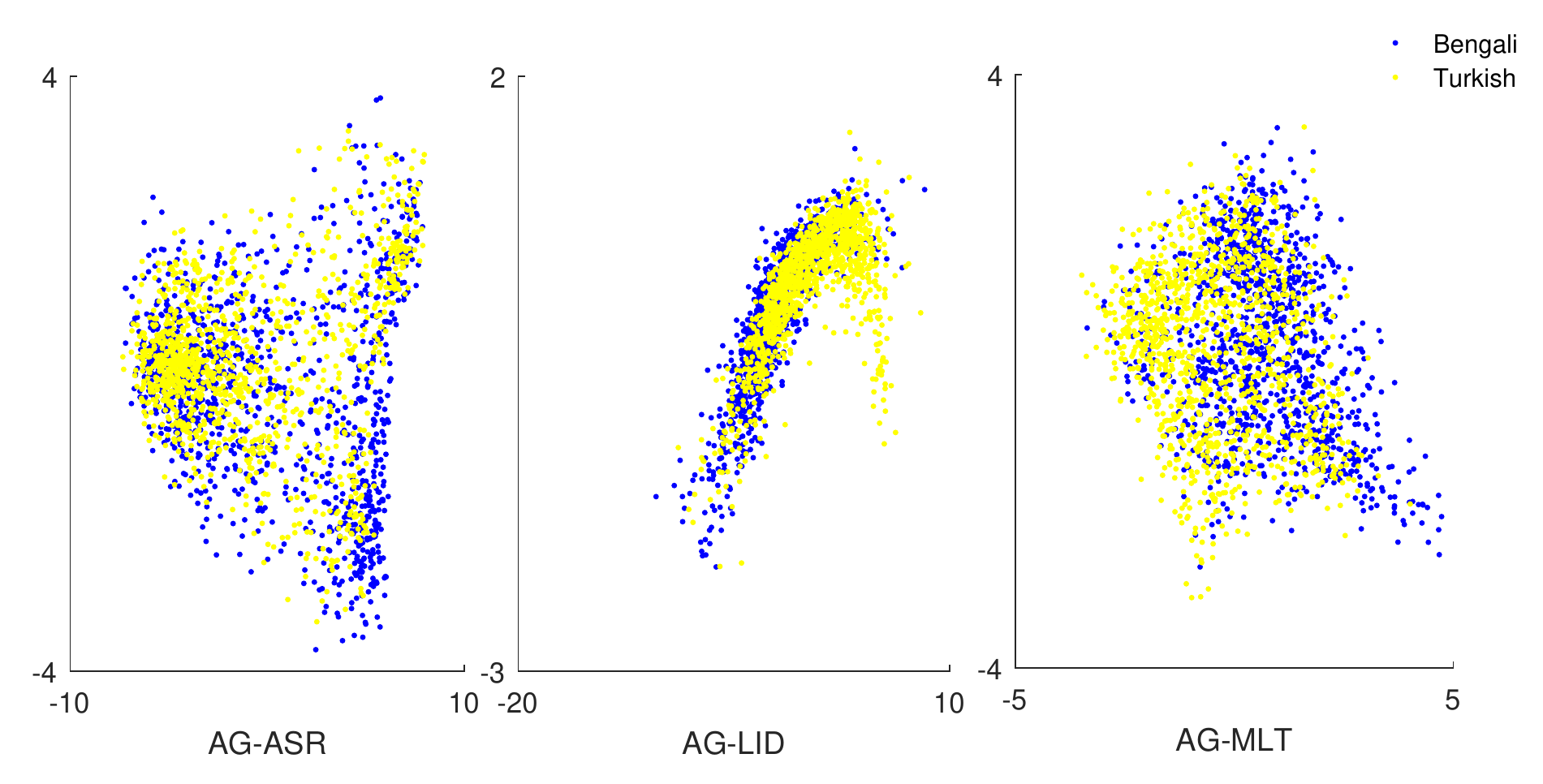}
  \caption{Features of the speech data from Bengali and Turkish, produced by the three phonetic RNNs.}
  \label{fig:visual34}
\end{figure}

\begin{figure}[h]
  \centering
  \includegraphics[width=\linewidth]{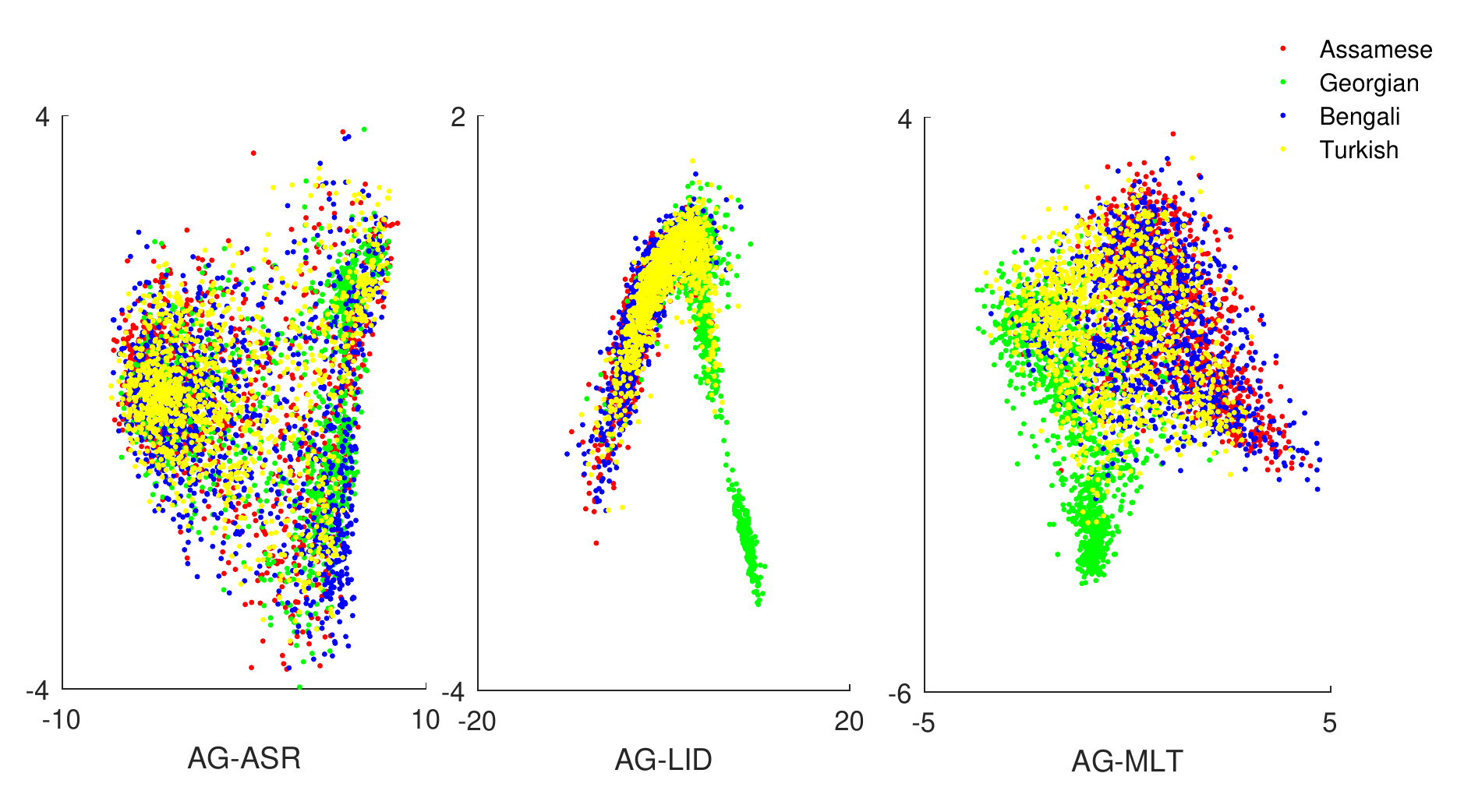}
  \caption{Features of the speech data from Assamese, Georgian, Bengali and Turkish, produced by the three phonetic RNNs}
  \label{fig:visual1234}
\end{figure}

The three baseline RNNs will be used as the candidates of the phonetic RNN. We visualize
the discriminative power of the phonetic features produced by these RNNs using PCA.
Specifically, $20$ test utterances are randomly selected from the test set for each language,
and these utterances are fed into the phonetic RNN frame by frame.
For each frame, the phonetic feature is read from the recurrent projection layer of
the tested RNN, and then is projected into the $2$-dimensional space by PCA.
Figure~\ref{fig:visual12} presents the distribution of the features for Assamese and Georgian,
the two languages `known' in the model training. Figure~\ref{fig:visual34} shows the distribution
for Bengali and Turkish, two language that are `unknown' in the model training.
Figure~\ref{fig:visual1234} shows the distribution of the features of all the four languages.
It can be observed that all the three RNNs possess certain discriminative capability for
both the known and unknown languages. Comparing the three models, the
features generated by the ASR-based RNN is clearly worse, and the features
generated by the multi-task RNN looks more discriminative.
Note that the phonetic features of Assamese and Bengali are highly overlapped, no matter
which phonetic RNN is used. This means that the four-language LID task will be
highly difficult, as we will see shortly.

\subsection{Phone-aware LID on known languages}

Due to the clear advantage of the AG-MLT in language discrimination,
we first choose this model to be the candidate of the phonetic RNN to
produce phonetic features. The LID RNNs are trained to discriminate the two known languages: Assamese and Georgian.
The results are shown in Table~\ref{tab:phone-lid-known}, where four configurations for the
`receiver' of the phonetic feature are tested: the input gate, the forget gate, the output gate and the $g$ function.
Compared to the results with the baseline RNNs (Table~\ref{tab:base}), introducing the phonetic feature
leads to clear performance improvement, on both the frame-level and the utterance-level, in terms of
both $C_{avg}$ and $EER$.

Then we use the AG-ASR and AG-LID models as the phonetic RNNs with only the best configurations above,
that is, the $g$ function or output gate as the receiver.
The AG-ASR results in better performance than both the baseline and AG-LID,
which further confirms our conjecture that phonetic information is valuable
for neural-based LID.

\begin{table}[thb!]
  \caption{\label{tab:phone-lid-known}{Results of phone-aware LID on known languages (Assamese and Georgian).}}
  \centerline{
    \begin{tabular}{c|c|cc|cc}
      \hline
           &            & \multicolumn{2}{c}{$C_{avg}$} & \multicolumn{2}{|c}{EER\%}\\
      \hline
      Phonetic RNN & Inf. receiver      & Fr. & Utt. & Fr. & Utt. \\
      \hline
      AG-MLT       & input gate    & 0.0762      & 0.0298    &  7.99    & 3.25    \\
      AG-MLT       & forget gate   & 0.0765      & 0.0280    &  8.01    & 3.20    \\
      AG-MLT       & output gate   & 0.0752      & 0.0286    &  7.91    & 3.15   \\
      AG-MLT       & $g$ function  & \textbf{0.0726}      & \textbf{0.0264}    &  \textbf{7.57}    & \textbf{2.92}    \\
      \hline
      AG-ASR       & output gate   & 0.0801      & 0.0298    &  8.38    & 3.28    \\
      AG-ASR       & $g$ function  & 0.0795      & 0.0283    &  8.38    & 3.08    \\
      AG-LID       & output gate   & 0.1310      & 0.0638    &  13.67   & 7.00   \\
      AG-LID       & $g$ function  & 0.1279      & 0.0640    &  13.45   & 6.95   \\
      \hline
   \end{tabular}
  }
\end{table}

\subsection{Phone-aware LID on unknown languages}

We now test the generalizability of the phonetic feature. Specifically,
we use the feature to help discriminate two new languages, i.e., the
languages that are unknown during the phonetic RNN training, which are Bengali and Turkish
in our experiment. To test the gain with the phonetic feature, the LID RNN
trained with the two target languages, Bengali and Turkish, denoted by BT-LID,
is used as the baseline. For simplicity, we only test the scenario where the AG-MLT is used as the phonetic RNN,
and the $g$ function is used as the feature receiver.

The results are shown in Table~\ref{tab:phone-lid-unknown}. It can be seen that
although the phonetic RNN has no knowledge of the two target languages,
the phonetic feature it produces is still highly valuable for the LID task.
This is understandable as the phonetic units are often shared by human languages,
and so the phonetic information the phonetic RNN provides is generally
valuable. This in fact demonstrates that the phonetic RNN can be trained
very flexibly, by using speech data of any languages. This is particulary
interesting for languages with too little training data to obtain a reasonable
phonetic RNN.

\begin{table}[thb!]
  \caption{\label{tab:phone-lid-unknown}{Results of phone-aware LID on unknown languages (Bengali and Turkish).}}
  \centerline{
    \begin{tabular}{c|c|cc|cc}
      \hline
                 &      & \multicolumn{2}{c}{$C_{avg}$} & \multicolumn{2}{|c}{EER\%}\\
      \hline
      Phonetic RNN & LID RNN   & Fr. & Utt. & Fr. & Utt. \\
      \hline
         -          & BT-LID    & 0.1740      & 0.0813    & 17.47     & 8.65    \\
      AG-MLT        & BT-LID    & 0.0869      & 0.0194    & 8.70    & 2.15    \\
      \hline
   \end{tabular}
  }
\end{table}

\subsection{Phone-aware LID on four languages}

The final experiment tests the LID performance on all the four languages.
The baseline system is the RNN LID model trained with the data of the four languages,
denoted by AGBT-LID. For the phone-aware system, the AG-MLT is used to
produce the phonetic feature, and the $g$ function is used as the feature
receiver. The results are shown in Table~\ref{tab:phone-lid-all}. Again,
the performance is greatly improved by involving the phonetic feature.

\begin{table}[thb!]
  \caption{\label{tab:phone-lid-all}{Results of phone-aware LID on four languages (Assamese, Georgian, Bengali and Turkish).}}
  \centerline{
    \begin{tabular}{c|c|cc|cc}
      \hline
                 &      & \multicolumn{2}{c}{$C_{avg}$} & \multicolumn{2}{|c}{EER\%}\\
      \hline
      Phonetic RNN & LID RNN   & Fr. & Utt. & Fr. & Utt. \\
      \hline
         -          & AGBT-LID    & 0.2131      & 0.1499     & 21.34     &  15.75    \\
      AG-MLT        & AGBT-LID    & 0.1531      & 0.0833     & 15.43     &   9.27   \\
      \hline
   \end{tabular}
  }
\end{table}


\section{Conclusions}
\label{sec:con}

We presented a phone-aware LSTM-RNN model for language identification. Our argument is that phonetic information
is important for LID. This information has been successfully used in the historical phonetic models such as the
famous PRLM system, but it has been largely ignored by the present pure acoustic methods, either the i-vector model
or the pure neural model. Particularly with the LSTM-RNN model, the inherent power on modeling temporal
dynamics with this model has been largely wasted without phonetic information involved. The phone-aware
architecture we proposed in the paper employs a deep neural model to produce phonetic features and these features are
propagated to the vanilla LSTM-RNN LID system. Our experiments conducted on the data of four languages of the Babel corpus
demonstrated that the phone-aware model can dramatically improve performance of the LSTM-RNN LID system.
In the future, we will test the phone-aware approach on more languages and under more complex conditions.
Particularly, we expect that with the phonetic information, the RNN-based LID may be significantly improved
on long utterances, by providing the phonetic normalization.

\newpage
\bibliographystyle{IEEEtran}
\bibliography{lid}

\begin{thebibliography}{10}
\providecommand{\url}[1]{#1}
\csname url@samestyle\endcsname
\providecommand{\newblock}{\relax}
\providecommand{\bibinfo}[2]{#2}
\providecommand{\BIBentrySTDinterwordspacing}{\spaceskip=0pt\relax}
\providecommand{\BIBentryALTinterwordstretchfactor}{4}
\providecommand{\BIBentryALTinterwordspacing}{\spaceskip=\fontdimen2\font plus
\BIBentryALTinterwordstretchfactor\fontdimen3\font minus
  \fontdimen4\font\relax}
\providecommand{\BIBforeignlanguage}[2]{{%
\expandafter\ifx\csname l@#1\endcsname\relax
\typeout{** WARNING: IEEEtran.bst: No hyphenation pattern has been}%
\typeout{** loaded for the language `#1'. Using the pattern for}%
\typeout{** the default language instead.}%
\else
\language=\csname l@#1\endcsname
\fi
#2}}
\providecommand{\BIBdecl}{\relax}
\BIBdecl

\bibitem{lamel1994language}
L.~F. Lamel and J.-L. Gauvain, ``Language identification using phone-based
  acoustic likelihoods,'' in \emph{Acoustics, Speech, and Signal Processing,
  1994. ICASSP-94., 1994 IEEE International Conference on}, vol.~1.\hskip 1em
  plus 0.5em minus 0.4em\relax IEEE, 1994, pp. I--293.

\bibitem{zissman1996comparison}
M.~A. Zissman \emph{et~al.}, ``Comparison of four approaches to automatic
  language identification of telephone speech,'' \emph{IEEE Transactions on
  speech and audio processing}, vol.~4, no.~1, p.~31, 1996.

\bibitem{li2007vector}
H.~Li, B.~Ma, and C.-H. Lee, ``A vector space modeling approach to spoken
  language identification,'' \emph{IEEE Transactions on Audio, Speech, and
  Language Processing}, vol.~15, no.~1, pp. 271--284, 2007.

\bibitem{Najim2011lang}
N.~Dehak, A.-C. Pedro, D.~Reynolds, and R.~Dehak, ``Language recognition via
  i-vectors and dimensionality reduction,'' in \emph{Proceedings of the Annual
  Conference of International Speech Communication Association (INTERSPEECH)},
  2011, pp. 857--860.

\bibitem{martinez2011language}
D.~Mart{\i}nez, O.~Plchot, L.~Burget, O.~Glembek, and P.~Matejka, ``Language
  recognition in ivectors space,'' \emph{Proceedings of Interspeech, Firenze,
  Italy}, pp. 861--864, 2011.

\bibitem{lopez2014automatic}
I.~Lopez-Moreno, J.~Gonzalez-Dominguez, O.~Plchot, D.~Martinez,
  J.~Gonzalez-Rodriguez, and P.~Moreno, ``Automatic language identification
  using deep neural networks,'' in \emph{Acoustics, Speech and Signal
  Processing (ICASSP), 2014 IEEE International Conference on}.\hskip 1em plus
  0.5em minus 0.4em\relax IEEE, 2014, pp. 5337--5341.

\bibitem{gonzalez2014automatic}
J.~Gonzalez-Dominguez, I.~Lopez-Moreno, H.~Sak, J.~Gonzalez-Rodriguez, and
  P.~J. Moreno, ``Automatic language identification using long short-term
  memory recurrent neural networks.'' in \emph{Interspeech}, 2014, pp.
  2155--2159.

\bibitem{gelly2016divide}
G.~Gelly, J.-L. Gauvain, V.~Le, and A.~Messaoudi, ``A divide-and-conquer
  approach for language identification based on recurrent neural networks,''
  \emph{Interspeech 2016}, pp. 3231--3235, 2016.

\bibitem{zazo2016language}
R.~Zazo, A.~Lozano-Diez, J.~Gonzalez-Dominguez, D.~T. Toledano, and
  J.~Gonzalez-Rodriguez, ``Language identification in short utterances using
  long short-term memory ({LSTM}) recurrent neural networks,'' \emph{PloS one},
  vol.~11, no.~1, p. e0146917, 2016.

\bibitem{lozano2015end}
A.~Lozano-Diez, R.~Zazo~Candil, J.~Gonz{\'a}lez~Dom{\'\i}nguez, D.~T. Toledano,
  and J.~Gonzalez-Rodriguez, ``An end-to-end approach to language
  identification in short utterances using convolutional neural networks,'' in
  \emph{Proceedings of the Annual Conference of the International Speech
  Communication Association, INTERSPEECH}.\hskip 1em plus 0.5em minus
  0.4em\relax International Speech and Communication Association, 2015.

\bibitem{jin2016lid}
M.~Jin, Y.~Song, I.~Mcloughlin, L.-R. Dai, and Z.-F. Ye, ``{LID}-senone
  extraction via deep neural networks for end-to-end language identification,''
  in \emph{Proc. of Odyssey}, 2016.

\bibitem{kotov2016language}
M.~Kotov and M.~Nastasenko, ``Language identification using time delay neural
  network d-vector on short utterances,'' in \emph{Speech and Computer: 18th
  International Conference, SPECOM 2016, Budapest, Hungary, August 23-27, 2016,
  Proceedings}, vol. 9811.\hskip 1em plus 0.5em minus 0.4em\relax Springer,
  2016, p. 443.

\bibitem{garcia2016stacked}
D.~Garcia-Romero and A.~McCree, ``Stacked long-term tdnn for spoken language
  recognition,'' \emph{Interspeech 2016}, pp. 3226--3230, 2016.

\bibitem{song2013vector}
Y.~Song, B.~Jiang, Y.~Bao, S.~Wei, and L.-R. Dai, ``I-vector representation
  based on bottleneck features for language identification,'' \emph{Electronics
  Letters}, vol.~49, no.~24, pp. 1569--1570, 2013.

\bibitem{ferrer2016study}
L.~Ferrer, Y.~Lei, M.~McLaren, and N.~Scheffer, ``Study of senone-based deep
  neural network approaches for spoken language recognition,'' \emph{IEEE/ACM
  Transactions on Audio, Speech and Language Processing (TASLP)}, vol.~24,
  no.~1, pp. 105--116, 2016.

\bibitem{tian2016investigation}
Y.~Tian, L.~He, Y.~Liu, and J.~Liu, ``Investigation of senone-based long-short
  term memory rnns for spoken language recognition,'' \emph{Odyssey 2016}, pp.
  89--93, 2016.

\bibitem{caruana1997multitask}
R.~Caruana, ``Multitask learning,'' \emph{Machine Learning}, vol.~28, no.~1,
  pp. 41--75, 1997.

\bibitem{matejka2006brno}
P.~Matejka, L.~Burget, P.~Schwarz, and J.~Cernocky, ``Brno university of
  technology system for nist 2005 language recognition evaluation,'' in
  \emph{Speaker and Language Recognition Workshop, 2006. IEEE Odyssey 2006:
  The}.\hskip 1em plus 0.5em minus 0.4em\relax IEEE, 2006, pp. 1--7.

\bibitem{salamea2016use}
C.~Salamea, L.~F. D'Haro, R.~de~C{\'o}rdoba, and R.~San-Segundo, ``On the use
  of phone-gram units in recurrent neural networks for language
  identification,'' \emph{Odyssey 2016}, pp. 117--123, 2016.

\bibitem{sak2014long}
H.~Sak, A.~W. Senior, and F.~Beaufays, ``Long short-term memory recurrent
  neural network architectures for large scale acoustic modeling,'' in
  \emph{Proceedings of the Annual Conference of International Speech
  Communication Association (INTERSPEECH)}, 2014, pp. 338--342.

\bibitem{povey2014parallel}
D.~Povey, X.~Zhang, and S.~Khudanpur, ``Parallel training of deep neural
  networks with natural gradient and parameter averaging,'' \emph{arXiv
  preprint arXiv:1410.7455}, 2014.

\bibitem{povey2011kaldi}
D.~Povey, A.~Ghoshal, G.~Boulianne, L.~Burget, O.~Glembek, N.~Goel,
  M.~Hannemann, P.~Motlicek, Y.~Qian, P.~Schwarz \emph{et~al.}, ``The kaldi
  speech recognition toolkit,'' in \emph{IEEE 2011 workshop on automatic speech
  recognition and understanding}, no. EPFL-CONF-192584.\hskip 1em plus 0.5em
  minus 0.4em\relax IEEE Signal Processing Society, 2011.

\end{thebibliography}

\end{document}